\documentclass[10pt,twocolumn,letterpaper]{article}

\usepackage{iccv}
\usepackage{times}
\usepackage{epsfig}
\usepackage{graphicx}
\usepackage{amsmath}
\usepackage{amssymb}
\usepackage{caption}

\usepackage[pagebackref=true,breaklinks=true,letterpaper=true,colorlinks,bookmarks=false]{hyperref}

\iccvfinalcopy %

\def\iccvPaperID{8034} %

\ificcvfinal\fi

\usepackage{color,xcolor}
\usepackage{epsfig}
\usepackage{graphicx}

\usepackage{comment}

\usepackage{pifont}

\usepackage{microtype}
\usepackage[font=small]{caption}
\usepackage{arydshln}
\usepackage{tabularx}
\usepackage{adjustbox}
\usepackage{array}
\usepackage{booktabs}
\usepackage{colortbl}
\usepackage{float,wrapfig}
\usepackage{hhline}
\usepackage{multirow}
\usepackage{subcaption} %
\usepackage[percent]{overpic}

\usepackage{stmaryrd}

\usepackage{amsmath,amsfonts,amssymb}
\usepackage{bm}
\usepackage{nicefrac}
\usepackage{microtype}
\usepackage{dsfont}
\usepackage{changepage}
\usepackage{extramarks}
\usepackage{fancyhdr}
\usepackage{setspace}
\usepackage{soul}
\usepackage{xspace}

\usepackage[pagebackref=true,breaklinks=true,colorlinks,bookmarks=false]{hyperref}
\usepackage{url}
\usepackage{xurl}

\usepackage{algorithm, algorithmic}
\usepackage{enumitem}
\usepackage[title]{appendix}

\usepackage{fp}
\usepackage{expl3}[2012-07-08]
\ExplSyntaxOn
\cs_new_eq:NN \fpeval \fp_eval:n
\ExplSyntaxOff

\usepackage{amsmath,amsfonts,bm}

\def\x{{x}}

\def\xi{{\x_i}}

\newcommand{\ignorethis}[1]{}

\newcommand{\myparagraph}[1]{\smallskip \noindent \textbf{#1}}

\def\eqref#1{equation~\ref{#1}}

\def\1{\bm{1}}

\DeclareMathAlphabet{\mathsfit}{\encodingdefault}{\sfdefault}{m}{sl}
\SetMathAlphabet{\mathsfit}{bold}{\encodingdefault}{\sfdefault}{bx}{n}

\DeclareMathOperator*{\argmax}{arg\,max}
\DeclareMathOperator*{\argmin}{arg\,min}

\newcolumntype{L}[1]{>{\raggedright\let\newline\\\arraybackslash\hspace{0pt}}m{#1}}
\newcolumntype{C}[1]{>{\centering\let\newline\\\arraybackslash\hspace{0pt}}m{#1}}
\newcolumntype{R}[1]{>{\raggedleft\let\newline\\\arraybackslash\hspace{0pt}}m{#1}}

\newcommand{\ignore}[1]{}

\makeatletter
\DeclareRobustCommand\onedot{\futurelet\@let@token\@onedot}
\def\@onedot{\ifx\@let@token.\else.\null\fi\xspace}

\makeatother

\definecolor{MyDarkBlue}{rgb}{0,0.08,1}
\definecolor{MyDarkGreen}{rgb}{0.02,0.6,0.02}
\definecolor{MyDarkRed}{rgb}{0.8,0.02,0.02}
\definecolor{MyDarkOrange}{rgb}{0.40,0.2,0.02}
\definecolor{MyPurple}{RGB}{111,0,255}
\definecolor{MyRed}{rgb}{1.0,0.0,0.0}
\definecolor{MyGold}{rgb}{0.75,0.6,0.12}
\definecolor{MyDarkgray}{rgb}{0.66, 0.66, 0.66}
\makeatletter
\let\@fnsymbol\@arabic
\makeatother
\begin{document}

\title{Paint by Word}

\author{%
Alex Andonian\footnotemark[1],
Sabrina Osmany\footnotemark[2],
Audrey Cui\footnotemark[1],
YeonHwan Park\footnotemark[1]\\
Ali Jahanian\footnotemark[1],
Antonio Torralba\footnotemark[1],
David Bau\footnotemark[1]
\\
\footnotemark[1]\, MIT CSAIL, \footnotemark[2]\, Harvard Graduate School of Design \\
{\tt\small \footnotemark[1] [andonian, audcui, parky, jahanian, torralba, davidbau]@csail.mit.edu} \\
{\tt\small \footnotemark[2] osmany@g.harvard.edu}
}

\let\oldtwocolumn\twocolumn
\renewcommand\twocolumn[1][]{%
    \oldtwocolumn[{#1}{
    \begin{center}%
\centering%
\includegraphics[width=\textwidth,trim=0 10pt 0 0]{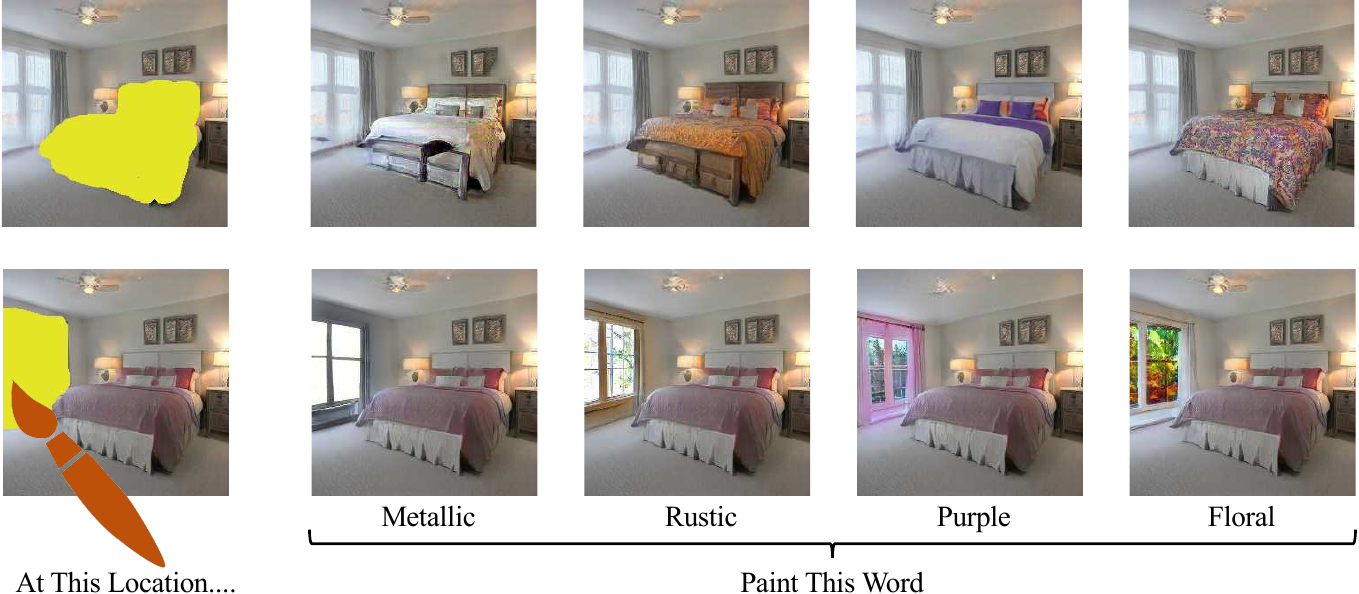}\\
\captionof{figure}{Paint by word.  Our method enables a user to create a scene by painting with words.  The user selects any visual concept using free text, and then applies that concept to an arbitrary region.  The mask and word are used together with a semantic similarity network to guide a generative model to create a realistic modification to the synthesized image.}
\label{fig:teaser}
\end{center}

    }]
}

\maketitle

\ificcvfinal\thispagestyle{empty}\fi
\begin{abstract}
We investigate the problem of zero-shot semantic image painting. Instead of painting modifications into an image using only concrete colors or a finite set of semantic concepts, we ask how to create semantic paint based on open full-text descriptions: our goal is to be able to point to a location in a synthesized image and apply an arbitrary new concept such as ``rustic'' or ``opulent'' or ``happy dog.'' To do this, our method combines a state-of-the art generative model of realistic images with a state-of-the-art text-image semantic similarity network. We find that, to make large changes, it is important to use non-gradient methods to explore latent space, and it is important to relax the computations of the GAN to target changes to a specific region. We conduct user studies to compare our methods to several baselines.
\end{abstract}

\begin{figure*}
\centering
\includegraphics[width=\textwidth]{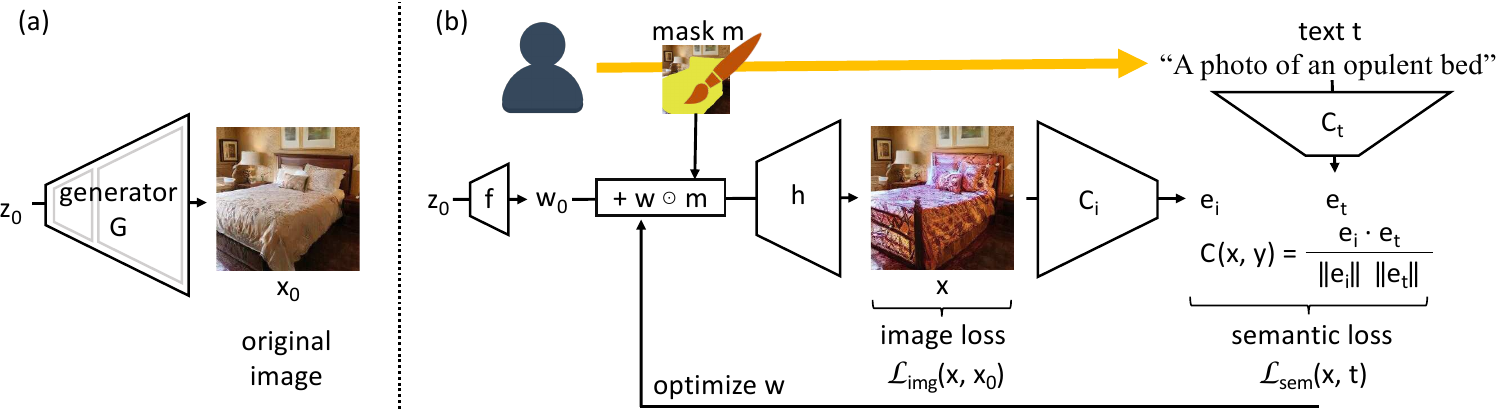}
\caption{An overview of our method.  (a) we begin with an image $x_0$ that is decoded from a latent $z_0$ by a pretranied generative model of realistic images $x_0 = G(z_0)$. (b) To edit the image, the user selects a region $m$ and gives an arbitrary run of descriptive text $t$.  The image generator $G(z) = h(f(z))$ is decomposed to expose a image representation $w$ that can be split spatially between the painted and unpainted region.  Then a text-image semantic similarity model $C(x, t)$ is used to define a semantic consistency loss that is used together with an image loss (minimizing changes outside the mask) to optimize $w$ inside the region to make the synthesized region match the given text.}
\label{fig:architecture}
\end{figure*}
\section{Introduction}

A writer can create vivid verbal imagery using just a few carefully-chosen words, but written scenes are abstract, without any specific physical form.
In this paper, we ask how to enable an artist to use words to build an image concretely, painting textually described visual concepts~\cite{osmany2023semiotics} at specific locations in a scene.  Our work is inspired and enabled by recent dramatic progress in text-to-image generation~\cite{razavi2019generating,radford2021learning}.  %
However, rather than having the AI compose the whole image, we ask how large models can be used collaboratively to let a person apply words as a paintbrush.  We wish to enable a user to paint brushstrokes on specific objects and regions within a generated image, then restyle, alter, or insert new objects by describing the desired visual concept with words.

The current work introduces the problem of zero-shot semantic image manipulation, in which the words and painting gestures available to the user are unconstrained and unknown ahead of time during model training.  Our goal is to enable a user to point at an image and apply an arbitrary new concept that may include visual concepts as wide-ranging as ``rustic'' or ``opulent'' or ``happy dog.''~\cite{osmany2023semiotics}

To paint with words, we pair large-scale generative adversarial networks (GANs~\cite{goodfellow2014generative}) that are trained unconditionally or with simple class conditioning~\cite{karras2020analyzing,brock2018large}, together with a full-text image retrieval network trained to measure the semantic similarity between text and an image~\cite{radford2021learning}.  We ask whether such a semantic similarity model can provide enough information to drive the image synthesis process to match a specific description within or beyond the domain on which the GAN is trained.   Then we ask how the structure of a GAN can be exploited to generate realistic images where a specific part of the image is modified to match a textual description.  We propose a simple architecture to enable painting with words, and we apply it using CLIP~\cite{razavi2019generating} paired with StyleGAN2~\cite{karras2020analyzing} and BigGAN~\cite{brock2018large}.

We conduct user studies to quantify the realism and accuracy of the changes made when using our method to edit a part of a generated image; we compare our method to several baselines, and we ask which types of visual concepts are easier or harder to paint by word. Finally we investigate several new semantic image manipulations that are enabled using our method.  Code and data will be available upon publication.

\section{Related Work}

Our application is inspired by recent progress in the text-to-image synthesis problem~\cite{ramesh2021zero,radford2021learning} as well as paint-based semantic image manipulation methods~\cite{bau2019semantic,park2019semantic,osmany2023semiotics}.

\vspace{3pt} \noindent \textbf{Text-to-image synthesis.} Synthesizing an image based on a text description is an ambitious problem that has attracted a variety of proposed solutions: initial RNN-based generators~\cite{gregor2015draw,mansimov2016generating} have been followed a series of by GAN-based~\cite{goodfellow2014generative} methods that have produced increasingly plausible images.  The GAN methods have adopted two distinct approaches to generating realistic images from text.  One is to generate images corresponding to text by sampling GAN latents that match semantics according to a separately trained image-text matching model~\cite{reed2016generative}; this idea can be refined by viewing the sampling as an energy-based modeling procedure~\cite{nguyen2017plug}.  The second approach is to train a conditional generator that explicitly takes a language embedding as input~\cite{reed2016generative}; the image quality of this approach can be improved using multi-stage generators~\cite{zhang2017stackgan,zhang2018stackgan++,zhang2018photographic}, and semantic consistency can be improved through careful design of architectures and training losses~\cite{xu2018attngan,zhu2019dm,qiao2019mirrorgan,yin2019semantics,li2019object,tao2020df}.  As an alternative approach, generating images from scene graphs instead of sentences~\cite{johnson2018image} can allow a generator to exploit logical structure explicitly.  Recent work stands apart by eschewing the use of GANs: the DALL-E~\cite{ramesh2021zero} model generates images from text autoregressively using a transformer~\cite{vaswani2017attention} based on GPT-3~\cite{brown2020language} to jointly generate natural text and image tokens using a VQ-VAE encoding~\cite{razavi2019generating}; outputs are sampled to maximize semantic consistency via a state-of-the-art image-text matching model CLIP~\cite{radford2021learning}.  Artist Murdock has observed that CLIP can be used as a source of gradients to guide a generator~\cite{murdock2021bigsleep,murdock2021aleph2image,murdock2021blog}.  The current paper is different from prior text-to-image work, because our goal is not to generate an image from text, but to define a paintbrush using text that enables a user to manipulate a semantics of a generated image at a specific painted location.

\vspace{3pt} \noindent \textbf{Semantic image painting.} Our work is also inspired by a series of methods that have enabled users to construct realistic images by painting a finite selection of chosen concepts at user-specified locations;  these can be thought of as ``paint by number.''  In this setting, there have again been two approaches enabled by GANs.  One approach is to find GAN latents to generate images that match a user's painted intention: this can be done to match color~\cite{zhu2016generative} or a finite vocabulary of visual concepts~\cite{bau2019gandissect,bau2019semantic,osmany2023semiotics} at a given location.  The second approach is to train an image-conditional Pix2Pix~\cite{isola2017image,wang2018high} model on the task of generating a realistic image with a semantic segmentation that matches a given painted input; the SPADE method \cite{park2019semantic} refines this approach to improve image quality given the limited information available in flat segmentation inputs.  While these methods all enable a user to paint images using a finite vocabulary of semantic concepts, the goal of the current work is to enable a user to paint an unlimited vocabulary of visual concepts, specified by free text.

\vspace{3pt} \noindent \textbf{GAN latent space methods.} Semantic manipulations in GAN latent spaces have been explored from several other perspectives.   Early work on GANs observed the presence of some latent vector directions corresponding to semantic concepts~\cite{radford2016unsupervised}, and further explorations have developed methods to identify interesting vector directions that steer latent space by using reconstruction losses~\cite{jahanian2019steerability,shen2020interpreting} or by learning from classification losses~\cite{denton2019detecting,goetschalckx2019ganalyze,osmany2023semiotics}. One disadvantage of this approach is that a semantic direction can only be found if we train a model to look for it in advance.  On the other hand, it has been observed that interior latents in GANs are partially disentangled~\cite{bau2019gandissect}.  So recent work has developed unsupervised methods for identifying disentangled interpretable directions in these interior latents~\cite{harkonen2020ganspace,wu2020stylespace}.  Our approach differs from the above because our method for identifying latent manipulations is supervised neither by a set of finite classes nor does it rely on disentanglement; rather it identifies semantic latent changes in a zero-shot manner using full-text image semantic similarity.

\vspace{3pt} \noindent \textbf{GAN inversion methods.}  We note that the leading state-of-the-art generative models are not trained with an encoder. Therefore, in order to apply GAN manipulation methods on real user-provided images, one must solve GAN inversion.  That is, we must be able to encode a given image in the latent space of the GAN.  Because several powerful editing methods are enabled by GAN latent manipulation, the problem of inverting a GAN is of widespread interest and is the subject of an active line of ongoing research~\cite{zhu2016generative,lipton2017precise,bau2019inverting,abdal2019image2stylegan,abdal2020image2stylegan,guan2020collaborative,richardson2020encoding,zhu2020indomain,zhu2020improved,huh2020ganprojection}.  The GAN inversion problem is complementary to our work.  In this paper, we shall assume that the image to be edited is represented in the latent space of the generator.

\section{Method}
\begin{figure}
\centering
\includegraphics[width=\columnwidth]{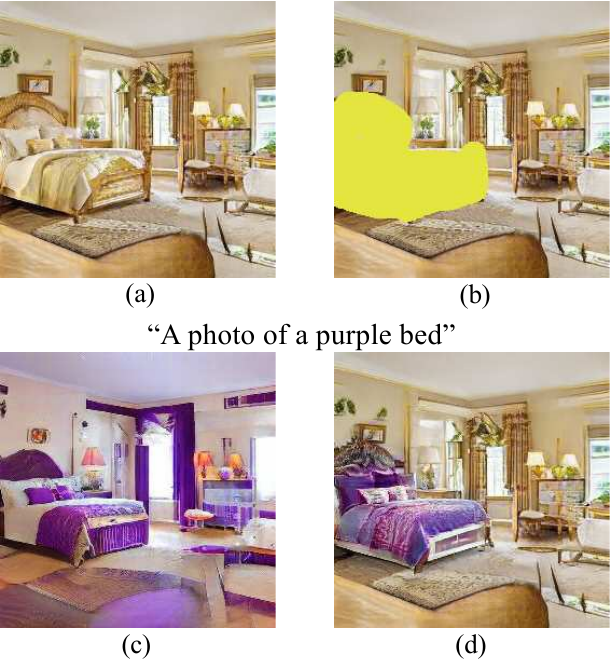}
\caption{Comparing the effect of optimizing with and without a spatially split generator.  The masked loss $C_{t,m}(x)$ is used to generate modifications of a generated image (a) that matches the description ``A photo of a purple bed.'' where the loss is masked to the user-painted region $m$ as shown in (b). The modification is done in two ways.  In (c), the original unmodified StyleGAN2~\cite{karras2020analyzing} model (trained on LSUN~\cite{yu2015lsun} bedrooms) is used and an optimizer finds a $G(z)$ that minimizes the masked loss.  In (d), the optimization is done over a spatially-split generator $G_{z, m}(w)$.  Although in both cases, the semantic loss examines only the region of the bed, only the split generator allows changes to be made in the user-specified region without altering the rest of the scene.
}
\label{fig:split-w-effect}
\end{figure}

There are two challenging problems that must be simultaneously solved in order to implement paint-by-word: first, we must achieve \emph{semantic consistency}, altering the painted part of the image to match the text description given by the user; and second, we must achieve \emph{realism}: the altered part of the image should have a plausible appearance.  In particular, the newly painted content should be consistent with the context of the rest of the image in terms of pose, color, lighting, and style.

We adopt a framework that addresses these two concerns using two separately trained networks: the first is a semantic similarity network $C(x, t)$ that scores the semantic consistency between an image $x$ and a text description $t$.  This network need not be concerned with the overall realism of the image.  The second is a convolutional generative network $G(z)$ that is trained to synthesize realistic images given a random $z$; this network enforces realism.  Given $G$ and $C$, we can formulate the following optimization to generate a realistic image $G(z^*)$ that matches descriptive text $t$:
\begin{align}
    z^* = \argmax_z C(G(z), t)
\label{eq:plug}
\end{align}
This simple approach factors the problem into two models that can be trained at large scale without any awareness of each other, and we shall use it as a starting point.
It allows us to take advantage of recent progress in state-of-the-art models that can be used for $G$ and $C$ (Such as StyleGAN~\cite{karras2020analyzing,karras2020training}, BigGAN~\cite{brock2018large}, CLIP~\cite{radford2021learning}, and ALIGN~\cite{jia2021scaling}).  Because our method allows the direct use of such large-scale models, this simple architecture is a promising approach for generating images from a text description even without any spatial conditions. We study this approach empirically in Section~\ref{sec:birds}.

\subsection{Semantic consistency in a region}

When providing semantic paint, the method of Equation~\ref{eq:plug} is not sufficient, because it does not direct changes specifically towards a user's chosen painted area. To focus effects on to one area, we direct the matching network $C$ to attend only to the region of the user's brushstroke instead of the whole image.  This can be done in a straightforward way: given a user-supplied mask $m$, we define a masked semantic similarity model
\begin{align}
C_{t,m}(x) = C(x \odot m, t)
\label{eq:masked-c}
\end{align}
Here $x \odot m$ is a projection that zeroes the components of $x$ outside the region of the mask $m$. 

By hiding the regions outside the mask from the semantic similarity model, the masked model $C_{t,m}(x)$ focuses the optimization on the selected region.  However, in practice we find that this is not enough to direct changes to only a single object.   Figure~\ref{fig:split-w-effect}(c) shows the effect of optimizing $\argmax_z C_{t,m}(G(z))$ to match a text description in a StyleGAN2~\cite{karras2020analyzing} model.  Although no gradients pass from $C$ to $G$ outside the masked region, $G$ has a computational structure that links the appearance of objects outside the mask to objects within the mask.

In order to allow a user to edit a single object, we must obtain a generative model that allows the region inside of the mask to be unlinked.  Interestingly, this can be done without training a new large-scale model from scratch.
\begin{figure}
\centering
\includegraphics[width=\columnwidth]{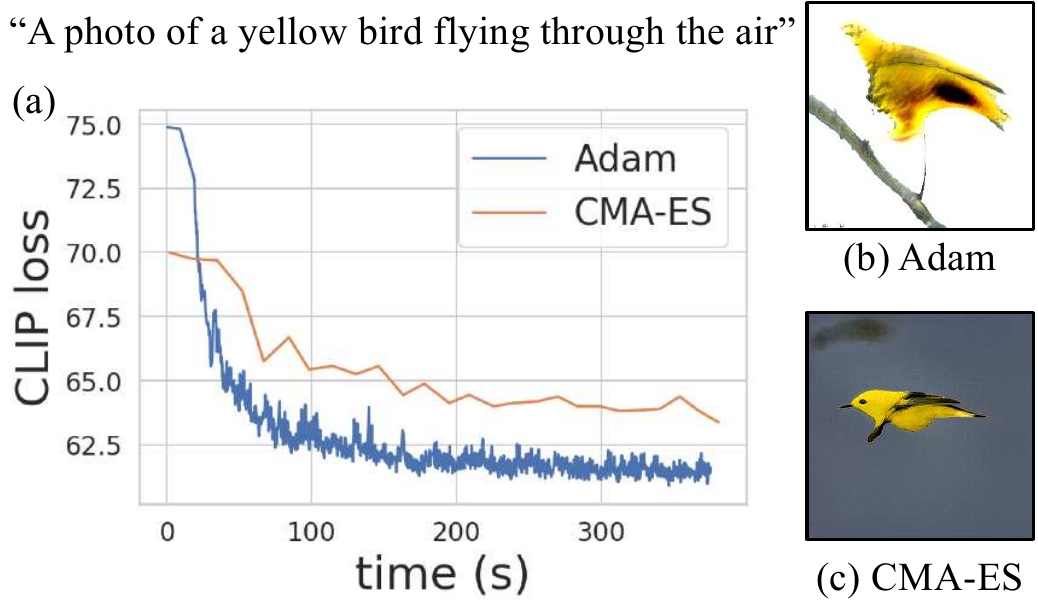}
\caption{Avoiding adversarial examples when optimizing an image to match the description ``A photo of a yellow bird flying through the air.''  The gradient descent method (Adam) obtains good scores while yielding an unrealistic image that looks neither like it is flying, nor like a real bird.  In practice, we obtain better results by using the non-gradient sampling method (CMA-ES). Although samples do not achieve scores that are as good, the generated samples avoid becoming adversarial, and user studies (Section~\ref{sec:birds}) show that results are both more realistic and accurate in practice.}
\label{fig:yellow-bird}
\end{figure}
\begin{figure*}
\centering
\includegraphics[width=\textwidth]{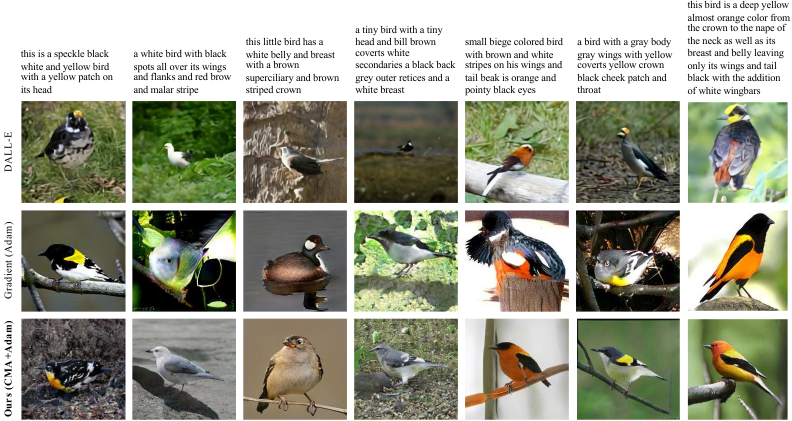} 
\caption{Qualitative comparison of a handful of birds from the user study conducted in Section~\ref{sec:birds}.  Our method (third row) is applied to synthesizing full images based on description of birds.  It is compared to an ablation where the Adam optimizer is used alone instead of CMA-ES; and we also compare our method to DALL-E~\cite{razavi2019generating}.  Note that images from our method tend to be more realistic more often when compared to the other implementations.}
\label{fig:bird-comparison}
\end{figure*}

\subsection{Generative modeling in a region}
To create a generative model that decouples the region inside and outside the mask, we exploit the convolutional structure of $G$.  We work with an intermediate latent representation of the image $w$ that has a spatial structure over the field of the image.  To access this structure, we express $G$ as two steps:
\begin{align}
x = G(z) & = h(f(z)) \\
w & = f(z)%
\end{align}
Given a mask $m$ provided by the user's brushstrokes, we then decompose $w$ into a portion outside the mask $w_0$ and a portion inside the mask $w_1$, as follows:
\begin{align}
    w &= w_0 + w_1 \\
    w_1 &= w \odot m
\end{align}
Here $w \odot m$ is a projection that zeroes the components of $w$ outside the region of the mask $m$.  In practice, we downsample $m$ to the relevant feature map resolution(s) of $w$, and use the Hadamard product to zero the outside components.

For a given original image $G(z)$ we can fix the original $w_0 = w - w\odot m$ while allowing $w_1$ to vary. This split gives us a new generative model, where the representation outside the mask $m$ is fixed as $w_0$, while the representation inside the mask is parameterized by $w$:
\begin{align}
G_{z,m}(w) = h(w_0 +  w \odot m)
\label{eq:masked-g}\end{align}
Splitting $w_0$ and $w$ in this way relaxes the problem, and allows $G_{z,m}(w)$ to generate images that the original model $G(z)$ could not generate.

In BigGAN, we use as $w$ the featuremap output of a layer of the network.  This featuremap has a direct natural spatial structure, and $w\odot m$ simply zeros the featuremap locations outside the mask.  In our BigGAN experiments we split the featuremap output of the first convolutional block of the generative network.

In StyleGAN, the natural interior latent, the $w$ vector, modulates featuremaps by changing channel normalizations uniformly across the spatial extent of the featuremap at all layers.  In our formulation, we split the $w$ style latent vector spatially by applying a one style modulation outside the user-specified mask, fixed as $w_0$, and another style modulation $w$ inside the mask.  In StyleGAN, the style modulation is applied across all layers, so we apply this split at the appropriate resolution at every layer of the model: the effect is to have a model that has two style vectors $w_0$ and $w$, instead of one.

This additional flexibility is instrumental for allowing the generator to create user-specified attributes in the region of interest without changing attributes of unrelated parts of the image.  Figure~\ref{fig:split-w-effect}(d) shows the effect of splitting the model: the appearance of one object can be changed without changing the appearance of other objects in the scene.  Combining (\ref{eq:masked-c}) with (\ref{eq:masked-g}) allows us to define the following masked semantic consistency loss in the region
\begin{align}
    \mathcal{L}_{\text{sem}}(w) = -  C_{t,m}(G_{z,m}(w))
\end{align}

In order to explicitly limit changes in the synthesized image outside the painited region, we apply the following image consistency loss:
\begin{align}
    \mathcal{L}_{\text{img}}(w) = d(x \odot (1-m), G_{z,m}(w) \odot (1-m))
\end{align}
Here $d$ denotes an image similarity metric: in our experiments, for $d$ we use a sum of an L2 pixel difference and the LPIPS~\cite{zhang2018perceptual} perceptual similarity.  In $\mathcal{L}_{\text{img}}$, the inverse of the user's masked region is applied, so this loss term does not limit changes inside the painted region.

The full loss adds the two loss terms:
\begin{align}
\mathcal{L}(w) &= \mathcal{L}_{\text{sem}}(w) +  \lambda_{img}  \mathcal{L}_{\text{img}}(w) \\
w^* & = \argmin_w \mathcal{L}(w)
\end{align}

\begin{figure}
\centering
\includegraphics[width=\columnwidth]{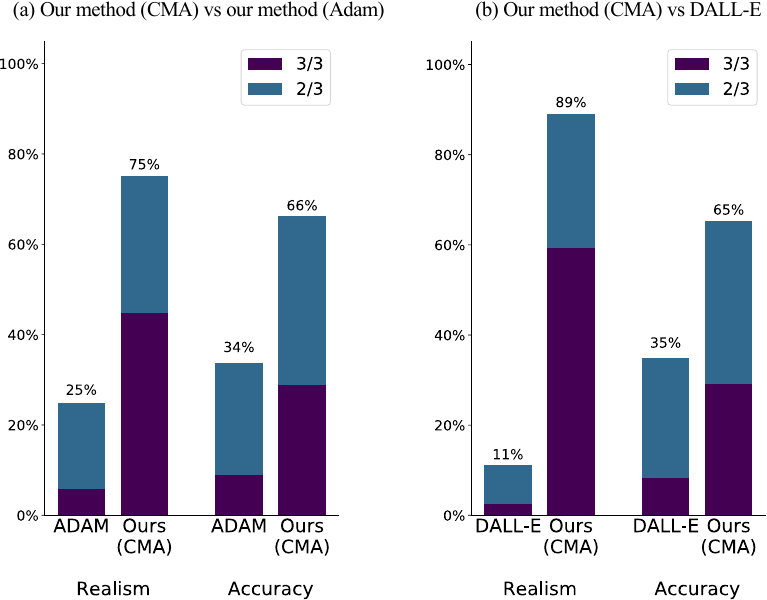}
\caption{User study comparing the semantic accuracy and realism of our basic method (Equation~\ref{eq:plug}).  We ask users to compare images generated by our method to two baselines; three workers examined each pair of images to make each assessment.   We compared our method as implemented using CMA~\cite{hansen2016cma} versus to our method using the Adam optimizer~\cite{kingma2015adam}.   Our CMA method outperforms our Adam method in both accuracy and realism.  To compare to a state-of-the-art baseline, we also compare our method to DALL-E~\cite{razavi2019generating}, which is text-to-image system that was trained on a much broader and more difficult domain.  Our method outperforms DALL-E in this setting, possibly because here our generator is trained to specialize on the narrow domain of birds.  In this test, our method pairs a 256-pixel StyleGAN2~\cite{karras2020analyzing,karras2020training} trained on CUB birds~\cite{wahCUB2011} with the CLIP~\cite{radford2021learning} \texttt{ViT/32B} text-image semantic similarity network.
}
\label{fig:birds-data}
\end{figure}

\subsection{Avoiding adversarial attack using CMA}

When using gradients to optimize a single image to a deep network objective, it is well-understood that it is easy to obtain an adversarial example~\cite{szegedy2013intriguing} that fools the network, achieving a strong score without being a typical representative of the distribution modeled by the network.  For example, it is easy to obtain an image that is classified as one class while having the appearance of an unrelated class.

The problem of adversarial attack also arises in our application.  Figure~\ref{fig:yellow-bird}(a) plots the loss when optimizing 
Equation~$\ref{eq:plug}$  using the Adam~\cite{kingma2015adam} optimizer, where $G$ is a StyleGAN2
~\cite{karras2020analyzing} trained on LSUN birds~\cite{yu2015lsun}, and where $C$ is the pretrained CLIP \texttt{ViT-B/32} network~\cite{radford2021learning}.  The convergence plot shows a rapid and stable-looking improvement as the image is modified to better match the phrase ``A photo of a yellow bird flying through the air,'' but the longer the optimization runs, the less like a flying bird the image becomes in practice.  Figure~\ref{fig:yellow-bird}(b) shows the image that results after several minutes of optimization: it achieves a strong score and yet has many artifacts that look obviously synthetic.  It does not resemble a photo of a bird.

One way to think about the problem is that our networks $C$ and $G$ were trained for optimal expected behavior over a distribution, not worst-case behavior for every individual.  So when optimizing a single instance it is not hard to find an individual point at which both models misbehave.

We solve this problem by switching optimization strategies.  Instead of seeking out a single optimal image, we use the Covariance Matrix Adaptation evolution strategy (CMA-ES)~\cite{hansen2016cma}, which is a non-gradient method that aims to optimize a Gaussian distribution to have minimal loss when random samples are drawn from the distribution.  Although CMA-ES may not achieve individual point losses that are as low as a gradient method like Adam, we find that in practice, the images that it obtains match modeled semantics very well, while remaining more realistic than the images produced by gradient descent: Figure~\ref{fig:yellow-bird}(c) shows a typical image from the distribution after running CMA optimization for an equivalent amount of time as Adam. In Section~\ref{sec:birds} we compare CMA-ES to Adam in a human evaluation.

\begin{figure*}
\centering
\includegraphics[width=\textwidth]{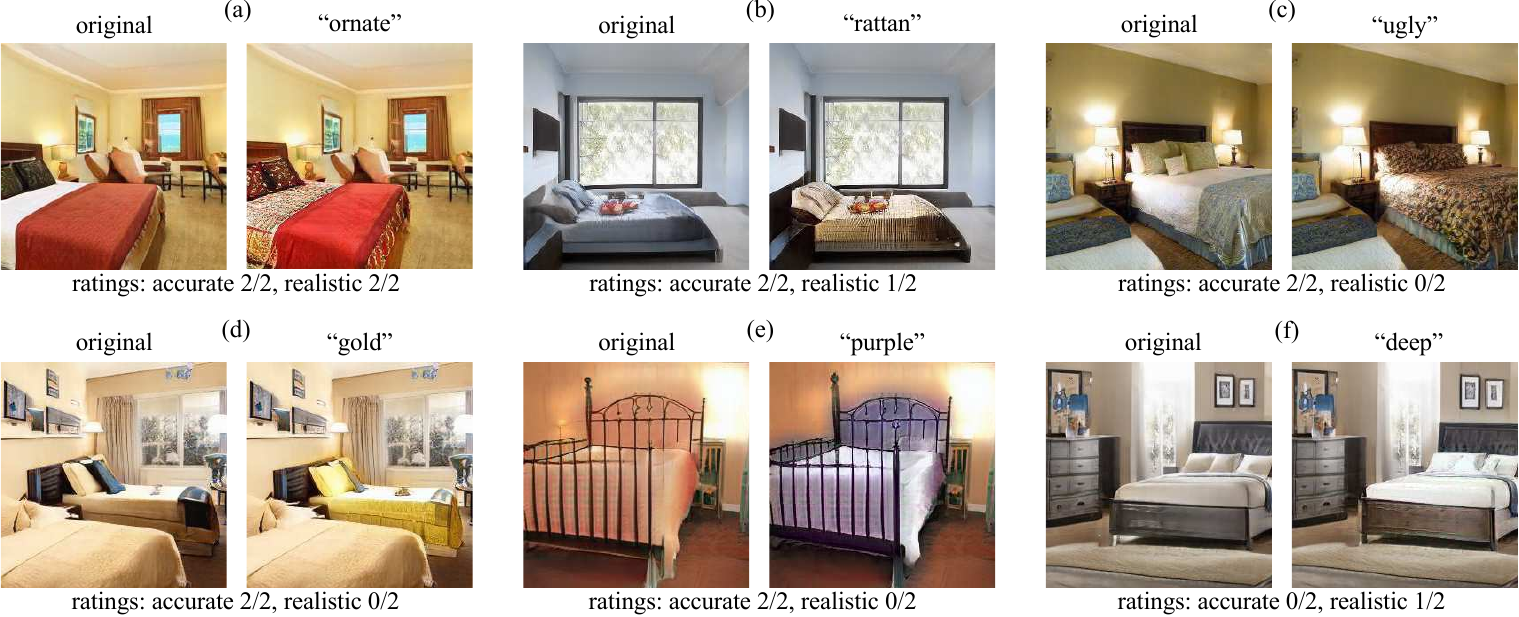} 
\caption{Representative successes and failures in a large-scale user study of image edits.  3000 edits were made on 300 randomly sampled StyleGAN2 bedroom outputs, and the edited image was compared to the original by crowdsourced workers. In cases a-e, 2/2 workers rated the modified image as a better match to the word than the original, and in case (f), both workers rated the modified image as a worse match to the word.  In case (a) 2/2 workers rated the edited image as more realistic than the original, in cases (b) and (f) the two workers split on realism, and in cases c-e, both workers rated the edited image as less realistic than the original.    }
\label{fig:user-study-cases}
\end{figure*}
\begin{figure}
\centering
\includegraphics[width=\columnwidth]{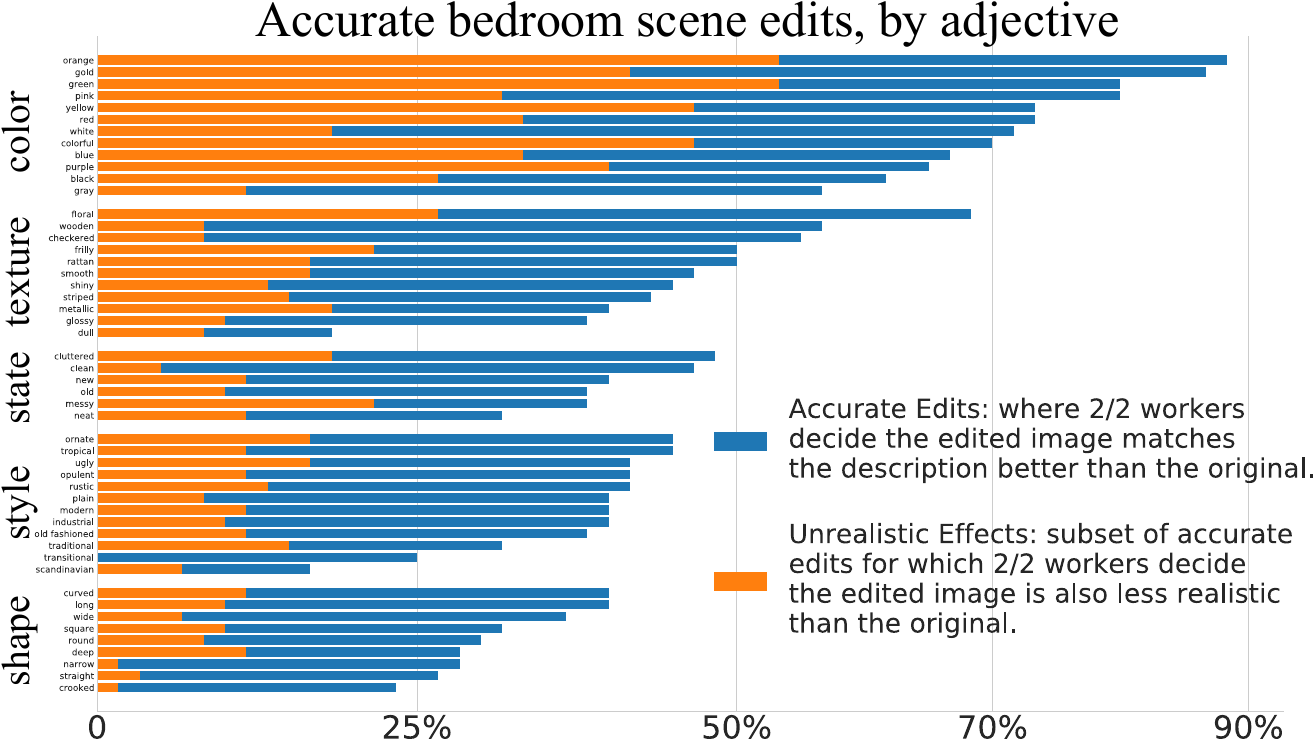}
\caption{A detailed breakdown of user edits that were considered accurate by 2/2 workers, according to individual word.  The most accurate edits among those we tested are for the color orange; and the least accurate are for the word "Scandinavian".  Edits that were considered less realistic than the original are shown in orange.}
\label{fig:semantic-categories}
\end{figure}
\section{Results}

To understand the strengths and weaknesses of our method, we conduct user studies in two problem settings.  Then we explore the types of new image manipulations that are enabled by paint-by-word.

\subsection{Testing full image generation}
\label{sec:birds}
Without a user-provided mask, our method is a simple factorization  of the text-conditional image generation problem (Equation~\ref{eq:plug}).  We wish to understand if this straightforward split of the problem has disadvantages compared to other approaches that explicitly condition image generation, so we begin by evaluating this architecture in comparison to a state-of-the-art model.

For $G$ we train a 256-pixel StyleGAN2~\cite{karras2020analyzing} with Adaptive Data Augmentation~\cite{karras2020training} on the CUB birds data set~\cite{wahCUB2011}. Standard training settings are used: the generator is unconditional, so no bird class labels nor text descriptions are revealed to the network during training.  For $C$ we use an off-the-shelf CLIP~\cite{radford2021learning} model, using the pretrained \texttt{ViT-B/32} weights; this model was trained on an extensive proprietary data set of 400 million image/caption pairs.  We generate images using two different techniques to maximize $C(G(z), t)$.   As a simple baseline we optimize $z$ using first-order gradient descent (Adam~\cite{kingma2015adam}).  Then we also optimize $z$ using CMA~\cite{hansen2016cma} followed by Adam.

To evaluate our method, we generate images from 500 descriptions of birds from a test dataset also used for evaluation of DALL-E in~\cite{razavi2019generating}.  To compare the images generated by our method to images generated by DALL-E, we conduct a user study on Amazon Mechanical Turk~\cite{barr2006ai} in which workers are asked to choose which of a pair of images are more realistic, and in a separately asked question, which of a pair more closely resemble a given text description.

The results are shown in Figure~\ref{fig:birds-data}.  We find that, for this test setting, our method gives competitive results.  The CMA optimization process is able to steer $G$ to generate an image that is found to be a better fit for the description than the DALL-E method 65.4\% of the time, and more realistic 89\% of the time.  The CMA method is also better than Adam alone, more accurate in 66.2\%; more realistic in 75.2\%.

This experiment is not intended to show that our simple method is superior to DALL-E: it is not.  Our network is trained only on birds; it utterly fails to draw any other type of subject.  Because of this narrow focus, it is unsurprising that it might be better at drawing realistic bird images than the DALL-E model, which is trained on a far broader variety of unconstrained images.  Nevertheless, this experiment demonstrates that it is possible to obtain state-of-the-art semantic consistency, at least within a narrow image domain, without explicitly training the generator to take information about the textual concept as input.

\subsection{A large scale user study of bedroom image edits}
\begin{table}
    \centering \small
    \begin{tabular}{l c c c c c}
    & color & texture & state & style & shape \\ \hline
    accurate (2/2) & 72.8 & 46.5 & 40.6 & 37.2 & 31.7 \\
    not accurate (0/2) & ~~3.9 & 13.2 & 20.0 & 20.4 & 22.8 \\ \hline
    realistic (2/2) & 13.1 & 18.2 & 21.7 & 26.1 & 25.4 \\
    not realistic (0/2) & 48.8 & 34.1 & 30.0 & 27.8 & 27.2
    \end{tabular}
    \caption{User study of edits, by semantic category.  Percentages shown include only cases where both evaluations agree.  ``Accurate'' counts cases where the edited image is a better match for the text.  ``Realistic'' counts cases where the edited image is more realistic than the original. In both cases the opposite counts are also shown.}
    \label{tab:bed-study}
\end{table}
Next, we conduct a large-scale user study on localized editing of objects in bedroom scenes.  For this study, we sample 300 images generated by StyleGAN2~\cite{karras2020analyzing} trained on LSUN bedrooms~\cite{yu2015lsun} and manually paint a single bed in each image.  Then for each image we perform 10 paint-by-word tests by applying one of a set of 50 text descriptions of colors, textures, styles, states, and shapes.  Humans compare each of the 3000 edited images to the corresponding original image: they evaluate whether one image is better described by the text description than the other, and separately whether one image is more realistic than the other.  Each comparison is done without revealing which image is the original.  Each comparison is evaluated by two people.

Results are tabulated by category in Table~\ref{tab:bed-study}, and charted by word in Figure~\ref{fig:semantic-categories}.  Our method can decisively edit a range of colors and textures, but it is also effective for a range of other visual classes to a lesser degree.  In all tested categories, most visual words can have some positive effect discerned by raters, however our method is weakest at being able to alter the shapes of objects.  Examples of success and failure cases from the test are shown in Figure~\ref{fig:user-study-cases}.

In Table~\ref{tab:bed-study}, we note that ratings of realism of edited images are worst in the same categories where the strength of the semantic consistency is best.  Some of the loss of realism is due to visible artifacts such as the blurred colors seen in Figure~\ref{fig:user-study-cases}(e).  However, other cases without noticeable visual artifacts are rated as less realistic by raters as in Figure~\ref{fig:user-study-cases}(c,d), including many of the most interesting edits in the study.  In these cases, the new painted styles are noticeable and stand out as less realistic than the original because the painted object now has an atypical style for the context - such as a gold bed in a cream-colored room, or an ugly bed in a hotel room.  The portion of effective edits that are rated as unrealistic is shown as orange bars in Figure~\ref{fig:user-study-cases}.

\subsection{Demonstrating paint-by-word on a broader\\ diversity of image domains}

To demonstrate the usefulness of our method beyond the narrow domains of birds and bedrooms, we apply our method on BigGAN models~\cite{brock2018large}.  We apply two BigGAN models.  First, we apply a BigGAN model trained on MIT Places~\cite{zhou2014learning}; and then we apply a BigGAN model trained on Imagenet~\cite{deng2009imagenet}.
\begin{figure*}
\centering
\includegraphics[width=\textwidth]{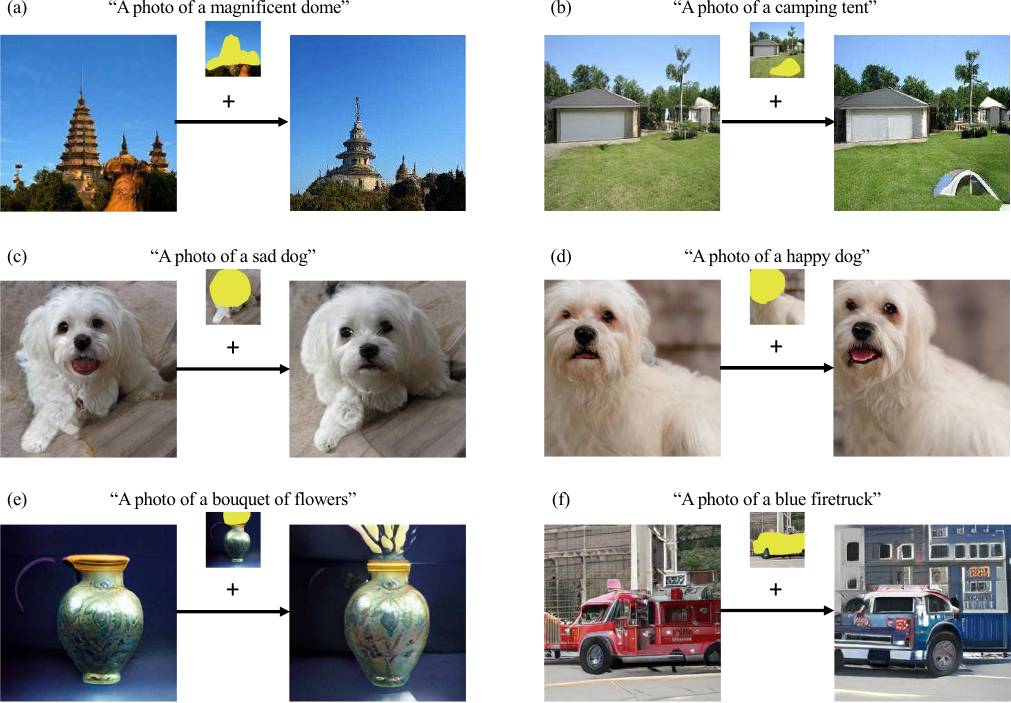}
\caption{Exploring edits using our method with BigGAN.  Because BigGAN models have been pretrained on very broad distributions, they provide an opportunity to experiment with a broad range of types of edits.}
\label{fig:biggan-edits}
\end{figure*}

Because the BigGAN models are trained on broad distributions of images, these generative models allow us to experiment with a wide variety of different editing operations.  In Figure~\ref{fig:biggan-edits}, We demonstrate our method's ability to alter the style of buildings outdoors (a); we also demonstrate the ability to add new objects that did not previously exist in the scene (b).  Both (a) and (b) demonstrate the potential of of BigGAN trained on Places to be used to manipulate general scene images.    On BigGAN Imagenet, we discover the ability to change whether a dog is happy or sad (c-d).   We can also add new objects to an appropriate context (e).  A challenging task is to force the model to synthesize objects outside its training domain.  Although the Imagenet training data contains no blue firetrucks, we can use language to specify that we wish for the model to render a firetruck as blue (f).  Using our method, it can create blue parts of a vehicle, but the out-of-domain task of creating a whole blue firetruck remains difficult and introduces distortions.

\section{Discussion}

We have introduced the problem of paint-by-word, a human-AI collaborative task which enables a user to edit a generated image by painting a semantic modification specified by any text description, to any location of the image.  Our work is enabled by the recent development of large-scale high-performance networks for realistic image generation and accurate text-image similarity matching.  We have shown that a natural combination of these powerful components can work well enough to enable paint-by-word.  Our user studies and demonstrations of effective edits have taken a first step in characterizing the opportunities, challenges and potential utility in this new application.

\medskip{\myparagraph{Acknowledgements.} We thank Aditya Ramesh at OpenAI for assistance with DALL-E evaluation data. We thank OpenAI, Google, and Nvidia for publishing weights for pretrained large-scale CLIP, BigGAN and StyleGAN models that make this work possible. We thank Hendrik Strobelt and Daksha Yadav for their insights, encouragement, and valuable discussions, and we are grateful for the support of DARPA XAI (FA8750-18-C-0004), and Signify Lighting Research.}

\paragraph{Author Contributions}

Alex Andonian contributed to developing the method, creating data sets, implementing and running experiments, analyzing results, and writing.
Sabrina Osmany developed the concept and prototype for abstract words like minimal, rustic, etc, including data sets, experiments, analysis.
Audrey Cui contributed to developing the method, implemented experiments.
YeonHwan Park contributed to implementing experiments, graphing and comparing results with other techniques. 
Ali Jahanian contributed to developing the method, implementing experiments, and writing.
David Bau contributed to developing the method, created data sets, implemented experiments, analyzed results, and writing.

{\small
\bibliographystyle{ieee_fullname}
\bibliography{egbib}
}

\end{document}